\documentclass[journal]{IEEEtran}

\usepackage{cite}
\usepackage{amsmath}
\usepackage{algorithmic}
\usepackage{url}
\usepackage{graphicx}
\usepackage{array}
\usepackage{amssymb}
\usepackage{multirow}
\usepackage[caption=false,font=footnotesize]{subfig}
\usepackage[pagebackref=true,breaklinks=true,colorlinks,bookmarks=false]{hyperref}
\usepackage{amsthm}
\usepackage{dsfont}
\usepackage{booktabs}
\usepackage{xcolor}
\usepackage[T1]{fontenc}
\usepackage{algorithmic}
\usepackage[ruled, vlined]{algorithm2e}
\usepackage{graphicx}
\usepackage{subfig}
\usepackage{booktabs}
\usepackage{threeparttable}

\SetKwInput{KwInput}{Input}             
\SetKwInput{KwOutput}{Output}
\graphicspath{{fig/}}
\def\eg{\emph{e.g.}}

\def\vs{\emph{vs.}}

\newcommand{\etal}{\textit{et al}.}
\begin{document}

\title{Attention Map Guided Transformer Pruning for Edge Device}

\author{
	Junzhu~Mao,
	Yazhou~Yao,
	Zeren~Sun,
	Xingguo~Huang,
	Fumin~Shen
	and~Heng-Tao~Shen

	\thanks{Junzhu Mao, Yazhou Yao  and Zeren Sun are with the School of Computer Science and Engineering, Nanjing University of Science and Technology, Nanjing 210094, China.}
	\thanks{Xingguo Huang is with the College of Instrumentation and Electrical Engineering, Jilin University, Changchun 130061, China.}
	\thanks{Heng-Tao Shen and Fumin Shen are with the School of Computer Science and Engineering, University of Electronic Science and Technology of China, Chengdu 611731, China.}	
}

\markboth{}%
{Shell \MakeLowercase{\textit{et al.}}: I2CRC}

\maketitle

\begin{abstract}

Due to its significant capability of modeling long-range dependencies, vision transformer (ViT) has achieved promising success in both holistic and occluded person re-identification (Re-ID) tasks. However, the inherent problems of transformers such as the huge computational cost and memory footprint are still two unsolved issues that will block the deployment of ViT based person Re-ID models on resource-limited edge devices. Our goal is to reduce both the inference complexity and model size without sacrificing the comparable accuracy on person Re-ID, especially for tasks with occlusion. To this end, we propose a novel attention map guided (AMG) transformer pruning method, which removes both redundant tokens and heads with the guidance of the attention map in a hardware-friendly way. We first calculate the entropy in the key dimension and sum it up for the whole map, and the corresponding head parameters of maps with high entropy will be removed for model size reduction. Then we combine the similarity and first-order gradients of key tokens along the query dimension for token importance estimation and remove redundant key and value tokens to further reduce the inference complexity. Comprehensive experiments on Occluded DukeMTMC and Market-1501 demonstrate the effectiveness of our proposals. For example, our proposed pruning strategy on ViT-Base enjoys \textup{\textbf{29.4\%}} \textup{\textbf{FLOPs}} savings with \textup{\textbf{0.2\%}} drop on Rank-1 and \textup{\textbf{0.4\%}} improvement on mAP, respectively. Code and models have been made available at \url{https://github.com/NUST-Machine-Intelligence-Laboratory/AMG}.

\end{abstract}


\ifCLASSOPTIONpeerreview
	\begin{center} \bfseries EDICS Category: 3-BBND \end{center}
\fi
\IEEEpeerreviewmaketitle

\section{Introduction}

\IEEEPARstart{R}{ecently}, vision transformer (ViT) \cite{dosovitskiy2020image} has attracted much attention and shed light for its' splendid performance on various computer vision applications. There is no doubt that person re-identification (Re-ID) is also included. Many researches have explored  ViT based Re-ID models and have achieved outstanding results on person Re-ID tasks for both occluded \cite{zhao2021incremental,wei2018glad,luo2020stnreid} and holistic \cite{zhao2020deep,zhao2021salience,jia2022learning} datasets. Compared with convolutional neural networks (CNNs), ViT has two advantages on Re-ID tasks, especially for occluded scenes. Firstly, multi-head self-attention (MSA) modules are introduced to capture long-range dependencies to learn more global structural features which are crucial for the model to focus on discriminative region \cite{li2021diverse}.  Secondly, transformer structure learns less inductive biases, hence has better absorbing capacity and generalization ability for sufficient training data \cite{dosovitskiy2020image, touvron2021training}. The powerful representations extracted by ViT can improve occluded Re-ID performance most straightforwardly. More researchers will devote themselves to exploring the potential of Re-ID methods based on ViT in the future.

\begin{figure}[t]
	\centering
	\includegraphics[width=0.98\linewidth]{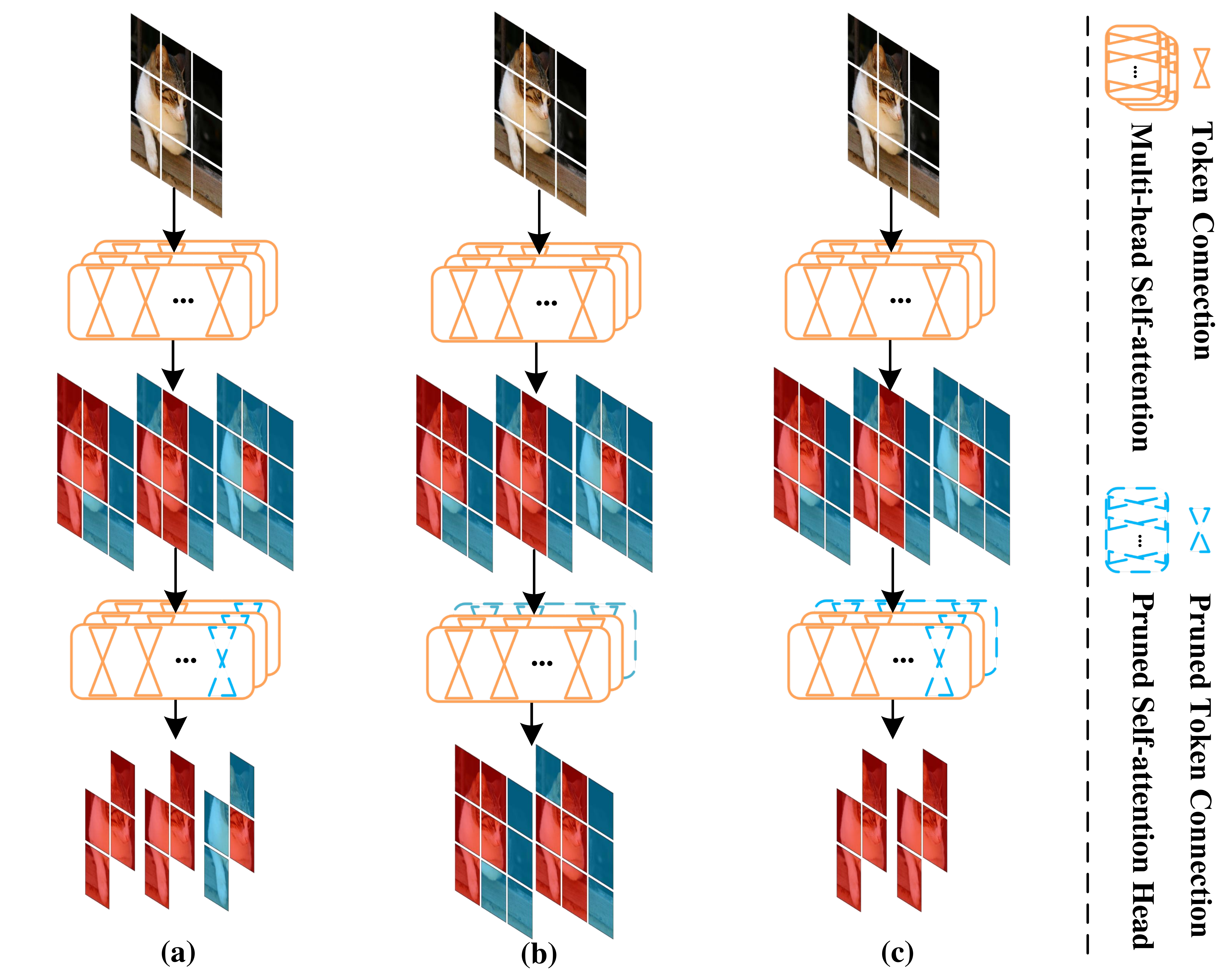}
	\caption{The comparison of (a) token pruning, (b) head pruning, and (c) our attention map guided pruning. Compared with (a) and (b), our proposed approach can obtain much more efficient pruning while retaining a comparable accuracy.}
	\label{fig1}
\end{figure}

Despite the promising performance of Re-ID methods with ViT embedded, the sequence-to-sequence structure of the transformer results in a higher computational cost as well as parameters. Those inherent problems will block the practical implementation of person Re-ID models on resource-limited edge devices for video surveillance~\cite{ wang2016joint,tan2017optimal,gaikwad2022end} and autonomous driving~\cite{camara2020pedestrian}. 
Pruning is a straightforward way to reduce the model size and computation cost for ViT, and it has been widely used in CNNs for efficient models~\cite{molchanov2016pruning, molchanov2019importance, chen2020lottery}. There are also many studies on ViT pruning, which can be summarized into the following two lines.
One line of works focus on token pruning (Fig.~\ref{fig1} (a)), which is specific to ViT whose input is a sequence of tokens~\cite{tang2021patch,wang2021pyramid,rethinking2021,xu2021evo}. 
These methods can effectively save the computational cost. However, they cannot reduce the model size. Another line explores head pruning (Fig.~\ref{fig1} (b)) to alleviate the memory bottleneck~\cite{chen2021chasing,yu2022width,yu2022unified}. The key point of head pruning is the actual removing of parameters for compression on edge devices, while most methods only take zero masking operation without considering the dimension align problems. 

In this work, we present a novel ViT pruning scheme which is a combination of the aforementioned two lines and is designed hardware-friendly. As shown in Fig.~\ref{fig1} (c), we introduce attention map to guide the pruning of ViT model. The motivation is that the self-attention module is the unique and core part of transformers, and can provide meaningful information for pruning. Specifically, we treat the learned attention maps from two aspects to prune ViT model on two dimensions with different focuses. We consider the features along \textbf{key} dimensions as the probability distributions, and further calculate entropy to present the uncertainty of attention. We then remove maps with high uncertainty by eliminating the corresponding head parameters of the QKV projection. The out projection is also pruned for dimension alignment. Hence we can structurally prune the parameters, which is friendly to hardware because no special design is needed. Thus, we can discard tokens which have small global attention, and further trim down the inference memory and computational cost. In the pruning process of our methods, we also take a layer-weighted global pruning scheme which can automatically lead to the best structure of pruned ViT models.
The main contributions of this work can be summarized as follows:

(1) We design pruning methods for ViT-based person Re-ID models with the consideration of occlusion and edge device deployment. 

(2) We propose entropy of the attention map as the criterion for head pruning and gradient-weighted similarity information of the attention map for token pruning. 

(3) We take a layer-weighted global pruning scheme to find the most suitable pruned structure. Extensive experiments on different benchmarks and ViT architectures demonstrate the superiority of our proposed approach.

\section{Related Work}
\subsection{Occluded Person Re-Identification}
The aim of person re-identification is to retrieve interested persons across camera views. In recent years, re-ID has made great progress \cite{zhao2020deep,wei2018glad, zhao2021salience}, and applied in various scenarios, such as video surveillance, autonomous driving, and activity analysis \cite{gaikwad2022end, camara2020pedestrian}. After observing the device applied to crowded public places, Zhuo~\etal~\cite{zhuo2018occluded} propose the occluded person Re-ID topic. Different from holistic person Re-ID, the query set is constructed with occluded pedestrian images, while the training set and gallery set are still holistic. Recent two mainstream methods deal with occlusion problems with the assistance of pose estimation~\cite{he2020guided,he2019foreground} and human parsing~\cite{huang2020human, yu2021neighbourhood}. With the popularity of ViT, some works also introduce transformer into occluded person Re-ID models~\cite{wang2022feature,li2021diverse,jia2022learning,he2021transreid}. Li~\etal~\cite{li2021diverse} design a transformer encoder-decoder architecture for diverse part discovery to solve the occluded problem. 
TransReid~\cite{he2021transreid} takes full advantage of ViT's powerful feature extraction capability by aggregating globally enhanced local information. Considering the simplicity of structure and outstanding performance, we choose TransReid as the baseline model for our pruning method.

\begin{figure*}[t]
	\centering
	\includegraphics[width=1.0\linewidth]{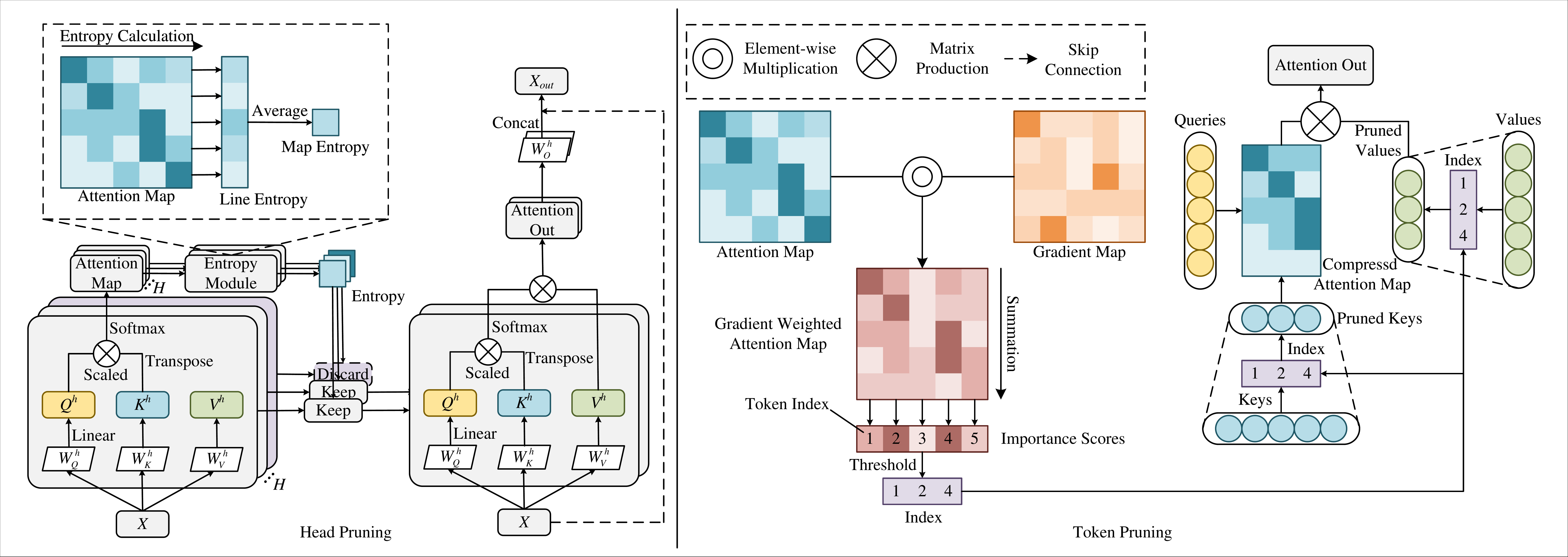}
	\caption{Framework. Our proposed attention map guided approach involves ViT pruning in two dimensions, including head and token pruning. For head pruning, we first train ViT model with a few iterations to obtain the generated attention maps. Then we calculate the average value of entropy along the key token direction for each row to get the entire attention map. We use the magnitude of entropy (visualized by the shade of color) to guide the pruning of heads, precisely, heads with large entropy will be removed. It should be noted that once head $\mathrm{h}$ are pruned, the corresponding weight parameters $\mathrm{W_Q^h, W_K^h, W_V^h, W_O^h}$ will also be eliminated, and the computation cost and memory footprint will also be reduced accordingly. For token pruning, the gradient-weighted attention maps are acquired by element-wise multiplication of the two prior obtained maps. The importance of each token (visualized by the shade of color) can be obtained by summing the gradient-weighted attention map along the direction of query tokens. We filter the indexes according to their importance, and the reserved indexes are leveraged to select the required key and value tokens for the attention calculation.}
	\label{fig2}
\end{figure*}

\subsection{Pruning for ViT} 
Recent ViT pruning work can be divided into two categories: reducing number of tokens and removing parameters. Tang~\etal~\cite{tang2021patch} develop a top-down patch slimming approach that removes redundant patches based on the reconstruction error of the pre-trained model.
Xu~\etal~\cite{xu2021evo} propose the structure preserving token selection and slow-fast updating strategies to fully utilize the complete spatial structure. These methods can effectively save the computational cost, but they cannot reduce the model size.  Zhu~\etal~\cite{zhu2021visual} introduce control coefficients and remove neurons with smaller ones to reduce the number of embedding dimensions, hence the model parameters can be compressed.
However, the unstructured pruning and zero masking operation are unfriendly for hardware efficiency which can provide limited acceleration and compression on edge devices.
Our structured head pruning with dimension align operation is designed for friendly edge device deployment. Our work aims to prune ViTs based on attention maps, which can afford guidance for retaining discriminative tokens for occluded Re-ID. Besides, we propose the entropy of the attention map as the prune criterion for head pruning rather than the commonly used Taylor criterion.

\section{The Proposed Approach}
In this work, we focus on the task of pruning ViT backbone of occluded person Re-ID model for obtaining hardware-friendly application on edge devices. Our proposed framework is illustrated in Fig.~\ref{fig2}. We prune on attention heads and tokens by exploring the information of attention maps.  


\subsection{Vision Transformer and Complexity Analysis}
\label{sec3.1}
The ViT model \cite{dosovitskiy2020image} splits input image to $\mathrm{N}$ patches and appends a class token, then feeds them to encoder transformer blocks which is kept same as the vanilla transformer \cite{vaswani2017attention}. The multi-head self-attention (MSA) and multi-layer perceptron (MLP) are the main components which cost most of the computation in the transformer block. We denote $\mathrm{Z}_{l-1} \in \mathbb{R}^{\mathrm{N \times d}}$ as the input of the $l$-th layer, then the attention calculation for head $\mathrm{h}$ can be formulated as:
\begin{equation}
\mathrm{Attn}(\mathrm{Z}_{l-1}, \mathrm{h}) = \mathrm{softmax}(\frac{\mathrm{Q}_{l}^{\mathrm{h}} \mathrm{K}_{l}^{\mathrm{{h}^T}}} {\sqrt{\mathrm{d}}}) \mathrm{V}_l^\mathrm{h},
\label{eq1}
\end{equation}
where $\mathrm{d}$ is the head embedding dimension. $\mathrm{Q}_{l}^{\mathrm{h}} = \mathrm{W}_l^{\mathrm{q, h}} \mathrm{Z}_{l-1}$, $\mathrm{K}_{l}^{\mathrm{{h}}}=\mathrm{W}_l^{\mathrm{k,h}}\mathrm{Z}_{l-1}$, and $\mathrm{V}_l^\mathrm{h}=\mathrm{W}_l^{\mathrm{v, h}}\mathrm{Z}_{l-1}$ are the query, key, and value of the $\mathrm{h}$-th head in the $l$-th layer, respectively. Thus, the formula for the MSA module can be written as:
\begin{equation}
\mathrm{MSA} (\mathrm{Z}_{l-1}) = \mathrm{Concat} [\mathrm{Attn}(\mathrm{Z}_{l-1}, \mathrm{h})]_{\mathrm{h}=1}^{\mathrm{H}}\mathrm{W}_l^\mathrm{o}.
\label{eq2}
\end{equation}
$\mathrm{H}$ is the number of heads. For $\mathrm{N}$ patches which are sent to the MSA module, we can calculate the complexity and parameters with Eq.~\eqref{eq1} and Eq.~\eqref{eq2}:

\begin{equation}
	\begin{split}
	\mathrm{C}(\mathrm{N, D, H, d}) & = \mathrm{C}((W^q,W^k,W^v) \times \mathrm{Z_{in}}) \\
	&~~~+ \mathrm{C}(\mathrm{Attn} \times W^o) + \mathrm{C}(\mathrm{Q \times K^{T}}) \times \mathrm{V}) \\
	& = 3\mathrm{NDHd} + \mathrm{NDHd} + \mathrm{2N^2Hd} \\
	& = \mathrm{2NHd}(\mathrm{2D + N}),
	\end{split}
	\label{eq3}
\end{equation}
and
\begin{equation}
\mathrm{P}(\mathrm{H, d}) = \mathrm{4DHd},
\label{eq4}
\end{equation}
The computation complexity is composed of three parts. The projection of input features $\mathrm{Z_{in}}$ to $\mathrm{Q,K,V}$, the projection of attention features $\mathrm{Attn}$ to output features $Z_{out}$, and the attention computation $(\mathrm{Q \times K^{T}}) \times \mathrm{V}$. The computational cost of the three parts are $3\mathrm{NDHd}$, $\mathrm{NDHd}$, and $\mathrm{2N^2Hd}$, respectively. $\mathrm{D}$ is the embedding dimension, and $\mathrm{D = H \cdot d}$ is true when the model is not pruned. We leverage this form to present the change in our pruning directly. We factor the formula to get the final expression. The computation of softmax and division are neglected. Parameters are the sum of four projection matrices $\mathrm{W^q, W^k, W^v, W^o}$.

\begin{figure}[t]
	\includegraphics[width=0.98\linewidth]{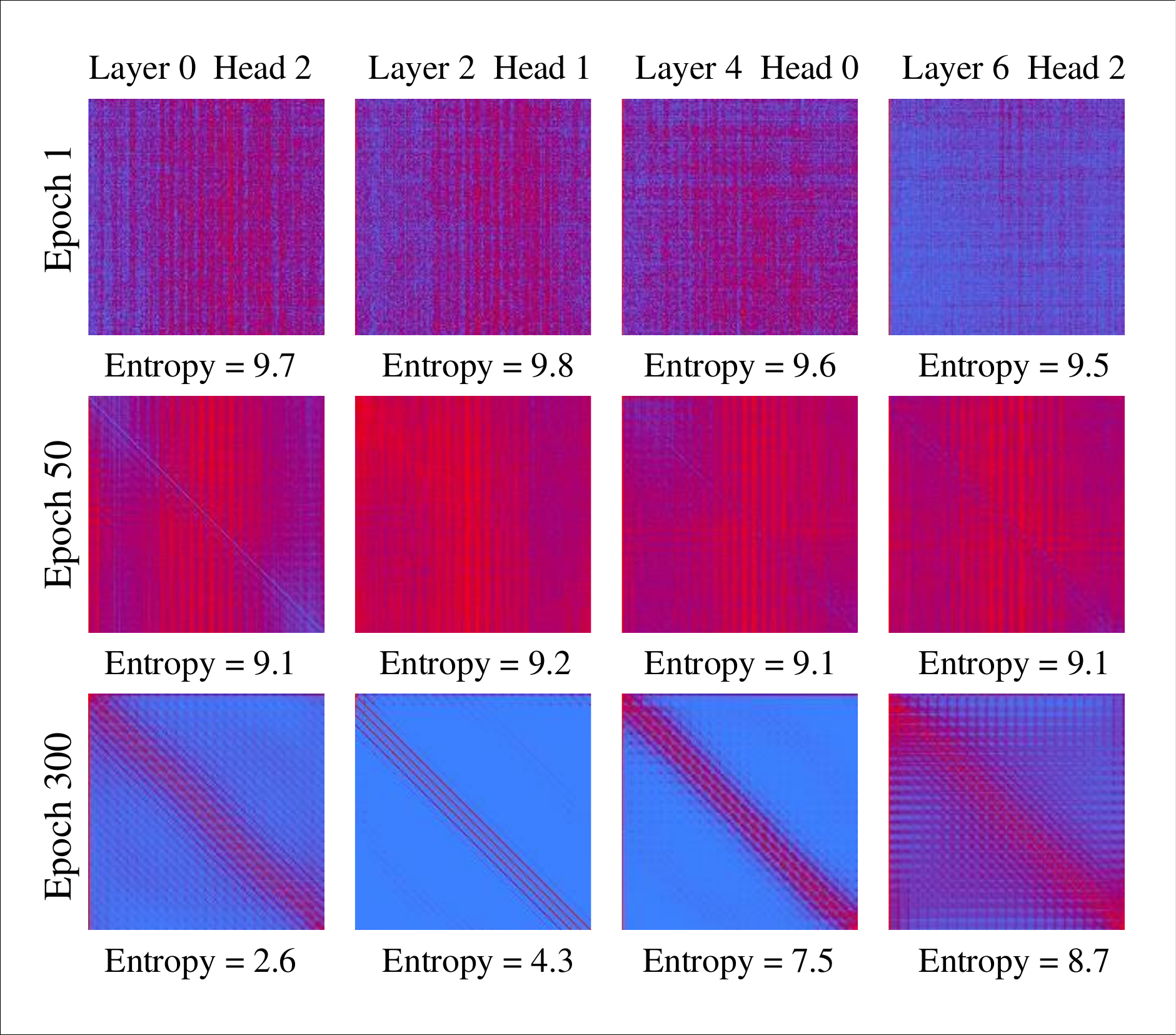}
	\vspace{-0.2cm}
	\caption{The attention maps in different training stage of ViT-Tiny model. Each map is the average result of one epoch. The red color indicates relatively high attention to all others while blue color means opposite.}
	\label{fig3}
\end{figure}


For our pruning methods, the number of attention heads and key-value tokens is pruned to ${\mathrm{H}}'$ and ${\mathrm{N}}'$, respectively. Then the Eq.~\eqref{eq3} and Eq.\eqref{eq4} can be rewritten as:
\begin{equation}
\mathrm{C}(\mathrm{N, N', D, H', d}) = \mathrm{2N'H'd(2D + N)},
\label{eq_cn}
\end{equation}
and
\begin{equation}
\mathrm{P}(\mathrm{D, H', d}) = \mathrm{4DH'd}.
\label{eq6}
\end{equation}
Eq.~\eqref{eq_cn} shows the computational cost of our pruned model. The formula indicates that with $\eta\%$ heads or tokens pruned, $\eta\%$ computation will be reduced in the MSA module. If both are pruned (heads and tokens), $\eta^2\%$ of the complexity of the MSA module will be reduced. From Eq.~\eqref{eq6}, the parameters of the MSA module are proportional to head numbers, while pruning tokens can't reduce the parameters.  

\subsection{Attention Map Guided Head Pruning}

\label{sec3.2}
Similar to the channel pruning of the convolution layer, head pruning is a structural parameter pruning mechanism. Thus, we can employ traditional channel pruning methods to prune the attention heads. Our initial idea is to leverage classical methods like L1 Norm \cite{li2016pruning} and Taylor Pruning \cite{molchanov2016pruning, molchanov2019importance}, but don't succeed in achieving better results. We consider the reason for our failure is that we haven't taken into consideration of the transformer's special characteristics.

The computation of the attention map is the core stage of the transformer block. To this end, we visualize the changes of the attention maps during the training from scratch process of the ViT model. The visualization results are present in Fig.~\ref{fig3}. All the attention maps showed are the mean result of one epoch training. We accumulate the attention maps and calculate the average for the common information learned by attention maps. 
By observing Fig.~\ref{fig3}, we can notice that in the initial stage of training, the model hasn't learned any useful information, so the attention maps are disordered and have a large entropy (\eg, epoch 1 in Fig.~\ref{fig3}). After a certain training epoch, the model learns some basic information and begins to present a certain pattern. However, attention to important information hasn't been learned, which means uniform attention and the attention maps still have large uncertainty. Therefore, the entropy is only reduced a little (\eg, epoch 50 in Fig.~\ref{fig3}). When the model finally converges, the entropy of all layers has reduced, which indicates the model has learned the information of the dataset. Important tokens are paid great attention, while unimportant tokens are paid fairly low attention. The shallow layers learn local attention and only focus on a few tokens. Hence, the entropy is significantly reduced (\eg, layer 0 and layer 2 at epoch 300 in Fig.~\ref{fig3}). In contrast, the deep layers tend to learn global attention, which needs to pay attention to more tokens. As a result, deep layers have higher entropy (\eg, layer 4 and layer 6 at epoch 300 in Fig.~\ref{fig3}).

Inspired by this, we can have the following conclusions. With the increase of useful information learned by the attention head, the model will pay more attention to some tokens, which makes the attention map more deterministic. On the contrary, when the attention head learns less information, it will have a unified focus on the overall situation, which results in great uncertainty. Information entropy can well measure the uncertainty of the attention map. In other words, the attention map with a large entropy value indicates that this head can only provide limited useful information and make little contribution to the overall situation, so it can be pruned.     

For attention map $\mathrm{A}^{\mathrm{h}, l}$ of layer $l$ and head $\mathrm{h}$, $\mathrm{A}_{i,j}^{\mathrm{h},l}$ denotes the $i$-th row and $j$-th column of attention map. In Eq.~\eqref{eq1}, a softmax operation is conducted along the key dimension after attention computation. It can be considered the similarity of $j$-th key token to the $i$-th query token in the original paper~\cite{vaswani2017attention}. We can also treat $\mathrm{A}_{i}^{\mathrm{h}, l}$ as a probability distribution for key tokens to the $i$-th query token mathematically. Hence the information entropy of attention map $\mathrm{A}^{\mathrm{h}, l}$ can be obtained by:
\begin{equation}
\mathrm{S}(\mathrm{A}^{\mathrm{h}, l}) = \sum_{i}^{\mathrm{N}}\sum_{j}^{\mathrm{N}}(-\mathrm{A}_{i,j}^{\mathrm{h}, l}\log \mathrm{A}_{i,j}^{\mathrm{h}, l}).
\label{eq7}
\end{equation}
According to the principle of maximum entropy~\cite{2021non,2021jo,pei2022eccv}, a uniform distribution of probability has maximum entropy. Therefore, we can prune the attention heads with a big entropy.

\subsection{Attention Map Guided Token Pruning}
\label{sec3.3}
The similarity calculation between query and key tokens is the main memory usage in a transformer block as it is proportional to the power of token length. To be specific, each key token needs to calculate similarity with all query tokens to obtain the attention map. However, not all tokens have a powerful contribution to the learning of global attention. In fact, a sparse token sequence may even lead to a better performance~\cite{rao2021dynamicvit} as images contain large regions of low-level texture and uninformative background. In the occlusion case, tokens with background and occlusion can be removed without impairing the model accuracy. 

By observing Fig.~\ref{fig3}, we can notice that the ViT model has learned a certain pattern on the attention map at epoch 300. The blue column of the visualized attention map means a small similarity between the corresponding key token to global query tokens. Small global similarity tokens typically make fewer contributions to the whole model and are less discriminative, thus we can prune these tokens which correspond to the blue columns. However, the simple summation of the similarity may neglect the global effect of the loss. Therefore, we propose to calculate the gradients of the attention map and perform element-wise multiplication for similarities and gradients.

Our token pruning strategy is inspired by Taylor Pruning \cite{molchanov2016pruning, molchanov2019importance}. The difference lies in that we compute the importance of the corresponding elements of key tokens on the attention map instead of simply calculating the importance of key tokens. For Taylor Pruning, the importance $\mathcal I$ of the $i$-th key token $k_i$ can be formulated as:    
\begin{equation}
\mathcal I(k_i) = \left | \frac{1}{\mathrm{D}}\sum_{\mathrm{d}}\frac{\delta \mathcal{L}}{\delta k_{i,\mathrm{d}}}k_{i,\mathrm{d}} \right |, 
\label{eq8}
\end{equation}
where $\frac{\delta \mathcal{L}}{\delta k_{i, \mathrm{d}}}$ is the gradient of loss function $\mathcal{L}$ with respect to the $\mathrm{d}$-th dimension of $k_i$. We compute the sum of the importance of key tokens corresponding to the column features on the attention map to replace the importance of key tokens. So our importance $\mathcal I$ of the $i$-th key token $k_i$ can be obtained by rewritten Eq.~\eqref{eq8} as: 
\begin{equation}
\mathcal I(k_i) = \left | \frac{1}{\mathrm{H}} \sum_{\mathrm{h}} \sum_{j}^{\mathrm{N}}\frac{\delta \mathcal{L}}{\delta \mathrm{A}_{i, j}^\mathrm{h}}\mathrm{A}_{i, j}^\mathrm{h} \right |. 
\label{eq9}
\end{equation}
$\mathrm{A}_{i, j}^\mathrm{h}$ denotes the $i$-th row and $j$-th column of the $\mathrm{h}$-th attention map. It's noticeable that we ignore the first token in calculating importance as the first token is the class token which shouldn't be pruned in the classification task. In addition, different from head pruning that we can remove the parameters to change numbers of head, token pruning has no relation to model structure. As shown in Fig.~\ref{fig2}, we add a index layer for pruning the key and value tokens.

\begin{table*}[t]
	\renewcommand\arraystretch{1.0}
	\centering
	\caption{The pruning results of our AMG method with TransReid-Base models on Market-1501 and Occluded-DukeMTMC. }
	\resizebox{17.4cm}{!}{
		\begin{tabular}{c|ccccccc}
			\toprule
			\textbf{Model}&\textbf{Head PR (\%)}&\textbf{Token PR (\%)} & \textbf{MSA Param. (M)} &\textbf{FLOPs (G)}&\textbf{Rank-1 (\%)}&\textbf{mAP (\%)}& \textbf{Throughput (img/s)}   \\ 
			\midrule
			\multicolumn{8}{c}{Market-1501} \\ 
			\midrule
			TR-Base       &           0      &   0  &  28.4   &           18.7   &           95.0     &          88.9   &      117               \\
			TR-AMG-Base-Token25&           0 &  25  &  28.4   & 14.5 (22.5\%$\downarrow$) &  94.8 (0.2$\downarrow$)&  88.5 (0.4$\downarrow$)  & 145 (1.24$\times$)  \\
			TR-AMG-Base-Token40&           0 &  40  &  28.4   & 12.8 (31.5\%$\downarrow$) &  94.8 (0.2$\downarrow$)& 87.8 (1.1$\downarrow$) &   161 (1.38$\times$) \\  
			TR-AMG-Base-Head25 &          25 &  0   &  21.3   & 16.2 (13.4\%$\downarrow$) &  95.0 (0.0$\uparrow$)  & 88.5 (0.4$\downarrow$) &   134 (1.15$\times$) \\
			TR-AMG-Base-Head40 &          40 &  0   &  17.0   & 14.6 (21.9\%$\downarrow$) &  94.5 (0.5$\downarrow$)& 88.1 (0.8$\downarrow$) &   142 (1.21$\times$) \\
			TR-AMG-Base-Head25-Token25 &  25 &  25  &  21.3   & 13.6 (27.3\%$\downarrow$) &  94.8 (0.2$\downarrow$)& 88.4 (0.5$\downarrow$)  &  155 (1.32$\times$) \\ 
			\midrule
			\multicolumn{8}{c}{Occlude-Duke} \\ 
			\midrule
			TR-Base       &           0      &   0  &  28.4   &           21.7   &           68.3     &          59.2   &      99               \\
			TR-AMG-Base-Token25&           0 &  25  &  28.4   & 16.9 (22.1\%$\downarrow$) &  67.9 (0.4$\downarrow$)& 59.4 (0.2$\uparrow$)  & 123 (1.24$\times$) \\
			TR-AMG-Base-Token40&           0 &  40  &  28.4   & 14.8 (31.8\%$\downarrow$) &  67.9 (0.4$\downarrow$)& 58.0 (1.2$\downarrow$)& 141 (1.42$\times$) \\  
			TR-AMG-Base-Head25 &          25 &  0   &  21.3   & 18.7 (13.8\%$\downarrow$) &  68.5 (0.2$\uparrow$)  & 59.7 (0.5$\uparrow$)  & 114 (1.15$\times$) \\
			TR-AMG-Base-Head40 &          40 &  0   &  17.0   & 16.7 (21.9\%$\downarrow$) &  66.3 (2.0$\downarrow$)& 58.2 (1.0$\downarrow$)& 120 (1.21$\times$) \\
			TR-AMG-Base-Head25-Token25 &  25 &  25  &  21.3   & 15.3 (29.4\%$\downarrow$) &  68.1 (0.2$\downarrow$)& 59.6 (0.4$\uparrow$)  & 137 (1.38$\times$) \\ 
			\bottomrule
	\end{tabular}}
	\label{tab_trans_b}	
\end{table*}

\begin{table*}[t]
	\renewcommand\arraystretch{1.0}
	\centering
	\caption{Exploring to minimize the size of TransReid on edge device while retaining the accuracy for occluded person Re-ID task.}
	\resizebox{17.4cm}{!}{
		\begin{tabular}{c|ccccccc}
			\toprule
			\textbf{Model}&\textbf{Head PR (\%)}&\textbf{Token PR (\%)}&\textbf{MSA Param. (M)} &\textbf{FLOPs (G)} & \textbf{Rank-1 (\%)} & \textbf{mAP (\%)} & \textbf{Troughput (img/s)} \\
			\midrule
			TR-Small            &        0           &      0       &      18.9     &      12.2               & 61.9                   & 52.4                  & 166  \\
			TR-AMG-Small-Token25&        0           &      25      &      18.9     & 9.5 (22.1\%$\downarrow$)& 62.2 (0.3$\uparrow$)   & 52.4 (0.0$\uparrow$)  & 204 (1.23$\times$)  \\
			TR-AMG-Small-Token40&        0           &      40      &      18.9     & 7.9 (35.2\%$\downarrow$)& 61.2 (0.7$\downarrow$) & 51.2 (1.0$\downarrow$)& 241 (1.45$\times$) \\
			TR-AMG-Small-Head25 &        25          &      0       &      14.2     &10.0 (18.0\%$\downarrow$)& 62.4 (0.5$\uparrow$)   & 52.7 (0.3$\uparrow$)  & 197 (1.19$\times$)\\
			TR-AMG-Small-Head40 &        40          &      0       &      11.3     &8.9  (27.0\%$\downarrow$)& 60.9 (1.0$\downarrow$) & 51.7 (0.7$\downarrow$)& 214 (1.29$\times$) \\
			TR-AMG-Small-Head67-Token40&        40   &      40      &      11.3     &6.8  (44.3\%$\downarrow$)& 61.9 (0.0$\downarrow$) & 50.2 (2.2$\downarrow$)& 276 (1.66$\times$)\\	
			\bottomrule
	\end{tabular}}
	\label{tab_trans_small}	
\end{table*}

\subsection{Layer Weighted Global Pruning Scheme and Fine-tuning}
Instead of the handcraft structure or the searched result in a large searching space, we employ the global pruning scheme to autonomously form the final structure of the pruned model. Compared with hand designing and heuristic researching, global pruning needs less prior experience and searching time. In addition, the pruning message can be fully utilized for finding the most suitable structure for the current task. Our global pruning scheme is simple to imply. First, we calculate the importance score of all the heads/tokens in the model. Then, we arrange them in descending order, removing the heads/tokens with the lowest score according to the pruning rate. 
When pruning the model in the global range, the difference between layers should be considered. As proposed in~\cite{tang2021patch}, deeper layers are less important for the redundant global attention. For better-allocated layer structures, we introduce a weight coefficient $\lambda$ for the calculation of importance score $\mathcal{I}$ in layer $l$. The formula is as follows:
\begin{equation}
 \mathcal{I}^{l}  = \left( 1 - \lambda l\right)\mathcal{I}^l
\end{equation}

\begin{table*}[t]
	\renewcommand\arraystretch{1.0}
	\centering
	\caption{ Pruning results on ImageNet and comparison with other ViT compression methods. $ \dagger $, $ \ddag $, and $ \S $ indicate a small, medium, and large computational cost, respectively.}
	\resizebox{17.4cm}{!}{
	\begin{tabular}{c|r|ccccc}
		\toprule
		\textbf{Types}         &             \textbf{Methods~~~~~~~~~} & \textbf{Publications} & \textbf{Image Size} & \textbf{Parameters} (M) &         \textbf{FLOPs} (G)          &     \textbf{Top-1 Acc.} (\%)     \\ \midrule
		\multirow{2}{*}{ $ \dagger $ } &  ViT-Tiny \cite{dosovitskiy2020image} &       ICLR2021        &       $384^2$       &            6            &                 4.6                 &               78.0               \\
		&         \textbf{ViT-AMG-Tiny-Token25} &           -           &       $384^2$       &            6            & \textbf{3.7 (19.6\%$\downarrow$)}  &  \textbf{78.4 (0.4$\uparrow$)}  \\ \midrule
		\multirow{3}{*}{\ddag}     & ViT-ResNAS-M \cite{liao2021searching} &       ICCV2021        &       $336^2$       &           97            &                15.2                 &               83.3               \\
		&   ViT-S16 \cite{dosovitskiy2020image} &       ICLR2021        &       $384^2$       &           22            &                15.3                 &               83.6               \\
		&                \textbf{ViT-AMG-Small-Head25} &           -    &       $384^2$       &           19            & \textbf{13.0 (15.0\%$\downarrow$)} & \textbf{83.5 (0.1$\downarrow$)} \\ \midrule
		\multirow{5}{*}{\S}       &  ViT-ResNAS-M\cite{liao2021searching} &       ICCV2021        &       $392^2$       &           97            &                15.2                 &               83.3               \\
		&     DeiT-B \cite{touvron2021training} &       ICML2021        &       $384^2$       &           86            &                55.4                 &               83.1               \\
		&             Swin-B \cite{liu2021swin} &       ICCV2021        &       $384^2$       &           88            &                47.0                 &               84.2               \\
		&   ViT-B16 \cite{dosovitskiy2020image} &       ICLR2021        &       $384^2$       &           86            &                55.4                 &               83.9               \\
		&        \textbf{ViT-AMG--Base-Token25} &           -           &       $384^2$       &           86            & \textbf{48.4 (12.6\%$\downarrow$)} &  \textbf{84.3 (0.4$\uparrow$)}  \\ \bottomrule
	\end{tabular}}
	\label{tab_imgnet}	
\end{table*}

We also add a soft distillation loss for better accuracy recovery when fine-tuning. The final loss function can be formulated as:
\begin{equation}
\mathcal{L} = \mathcal{L}_{CE} \left( y,p\right) + \alpha \mathcal{L}_{KL}\left( q,p\right)
\end{equation}
where $p$ is the output softmax probability of pruned model, $q$ is the corresponding value of original model, and $y$ is the true label. $\mathcal{L}_{CE}$ and $\mathcal{L}_{KL}$ denote the cross-entropy loss and the KL-divergence loss, respectively. $\alpha$ balances the weight of the teacher's knowledge.

\section{Experiments}
\subsection{Experimental Setup}
\paragraph{\textbf{Baseline Models}} We train TransReid~\cite{he2021transreid} as the baseline models for our AMG pruning method, 
which uses a pure transformer framework for the object ReID task with two proposed modules, the jigsaw patch module (JPM) and the side information embedding (SIE) module.  

\begin{table}[t]
	\setlength{\tabcolsep}{2.8mm}
	\renewcommand\arraystretch{1.02}
	\centering
	\caption{Comparisons with sota person Re-ID methods on both holistic and occluded datasets.}
	\begin{tabular}{{l}*{3}{c}}
		\toprule
		Methods                                   & Publication &    Rank-1     & mAP           \\ 
		\midrule
		\multicolumn{4}{c}{Occlude-Duke}                                                        \\ 
		\midrule
		PVPM~\cite{gao2020pose}                   &   CVPR20    &      47       & 37.7          \\
		PGFA~\cite{miao2019pose}                  &   ICCV19    &     51.4      & 37.3          \\
		HOReID~\cite{wang2020high}                &   CVPR20    &     55.1      & 43.8          \\
		OAMN~\cite{chen2021occlude}               &   ICCV21    &     62.6      & 46.1          \\
		PAT~\cite{li2021diverse}                  &   CVPR21    &     64.5      & 53.6          \\
		FED~\cite{wang2022feature}                &   CVPR22    &     68.1      & 56.4          \\
		TransReID~\cite{he2021transreid}          & ICCV21      &     66.4      & 59.2          \\
		\textbf{TR-AMG-Base-Head25}               &     -    & \textbf{68.5}    & \textbf{59.7} \\ 
		\midrule
		\multicolumn{4}{c}{Market-1501}                                                         \\ 
		\midrule
		PGFA~\cite{miao2019pose}                  &   ICCV19    &     91.2      & 76.8          \\
		PCB~\cite{sun2018beyond}                  &   ECCV18    &     92.3      & 77.4          \\
		OAMN~\cite{chen2021occlude}               &   ICCV21    &     92.3      & 86.3          \\
		BoT~\cite{luo2019bag}                     &   CVPRW19   &     94.1      & 85.7          \\
		PAT~\cite{li2021diverse}                  &   CVPR21    &  \textbf{95.4}& 88.0          \\
		TransReID~\cite{he2021transreid}          &   ICCV21    &     95.0      & \textbf{88.9} \\
		\textbf{TR-AMG-Base-Head25}               &-            & 95.0          & 88.5          \\ \bottomrule
	\end{tabular}
	\label{tab_compare_sota}	
\end{table}

\paragraph{\textbf{Datasets and Evaluation Metrics}}
We validate our approach on both holistic and occluded person Re-ID datasets. Market-1501~\cite{zheng2015scalable} is a widely used holistic person Re-ID dataset that contains 12,936 training images of 751 persons, 3,368 query images, and 19,732 gallery images of 750 persons. These images are captured by 6 cameras. Few persons are occluded in the dataset. Occluded-DukeMTMC~\cite{miao2019pose} is the most challenging occluded person Re-ID dataset with 15,618 training images of 702 persons, 2,210 query images of 519 persons, and 17,661 gallery images of 1,110 persons. We further conduct validation on ImageNet~\cite{deng2009imagenet} which is a large-scale dataset, consisting of 1.3 million training and 50,000 validation images with various spatial resolutions and 1000 classes. It is widely used as the benchmark for classification tasks and it is evaluated by the top-1 validation accuracy. We evaluate the performance of TransReid and its pruned model on Re-ID over two standard evaluation metrics: Cumulative Matching Characteristics (CMC) and Mean Average Precision (mAP). For the evaluation of model compression, we adopt the widely-used metrics, i.e., the number of parameters and required Float Points Operations (denoted as FLOPs). We also use the number of throughput images per second to show the improvement of our pruning methods directly. We show the number of attention parameters (MSA Param. for short) because we only prune the MSA module and keep the MLP module unchanged.

\paragraph{\textbf{Implementation Details}}
We first train the two baseline models on Market-1501 and Occluded DukeMTMC following the training scheme in the original paper, respectively. The trained models will be used as the base model for pruning, and also the teacher model for knowledge distillation. Then we prune the base models using our AMG method for both token and head pruning. The pruned model will be named like this: TR-AMG-Base-Token40-Head40. TR is the abbreviation for TransReid. Base means the base variant of ViT, we also use the Tiny and Small version, the details are shown on~\cite{touvron2021training}. The patch size is set to be $16\times 16$ and the input size $256 \times 128$ for TransReid if not specially mentioned. The number 40 following Token and Head indicates that 40\% of tokens and heads are pruned. 
For head pruning, we take an iterative strategy. Only part of the heads are pruned at one iteration. Different from head pruning, our token pruning method removes all the tokens that are considered redundant at one time. After pruning, the fine-tuning epochs are set to 30. For a fair comparison, we set the learning rate and weight decay as 0.0001 and 0.001 when fine-tuning, respectively. We also fine-tune the base model without pruning for 30 epochs to get abundant training epochs and choose the best results as the baseline accuracy. All the experiments are conducted on a single Nvidia Tesla V100 GPU with 16GB VRAM.

\subsection{Experimental results}

\paragraph{\textbf{Pruning Results on ViT-Base}} 
We prune TR-Base models on Market-1501 and Occluded-DukeMTMC to validate the performance of our pruning method. Table ~\ref{tab_trans_b} show the experimental results. By observing the results of four groups of experiments, we have the following observations. 
First, pruning on Occlude-Duke has more accuracy drop than on Market-1501. With the same 40\% of heads pruned for TransReid, we have 2.0\% and 1.0 \% drop of Rank-1 and mAP on Occluded-Duke, respectively. In contrast, on Market-1501, the Rank-1 and mAP only have a reduction of 0.5\% and 0.8\%, respectively. It aligns with our expectations because the Occlude-Duke dataset is much more challenging than the holistic dataset Market-1501. Hence, a larger accuracy drop will occur when the models are pruned to a moderate size. Finally, with a low pruning rate, we have nearly undamaged accuracy (\eg, drop 0.2\% of Rank-1 and 0.4\% of mAP when pruning 25\% tokens of TransReid on Market-1501) or even improved performance (0.2\% and 0.5\% increase of Rank-1 and mAP when removing 25\% heads of TransReid on Occluded-Duke). However, when the pruning rate grows too large, the accuracy will significantly decrease (2.0\% and 1.0 \% drop of Rank-1 and mAP on Occluded-Duke when 40\% heads of TransReid are eliminated). The reason is that the redundancy of heads and tokens is limited. When too many heads or tokens are pruned, even important ones will be removed. Therefore, we combine the head and token pruning to obtain a higher compression rate while retaining the accuracy (\eg, 29.4\% of FLOPs reduction and 1.37$\times$ speed up of TransReid model on Occlude-Duke with 0.2\% drop of Rank-1 and 0.4\% increase of mAP).

   \begin{table}[t]
	\renewcommand\arraystretch{1.0}
	\caption{The comparisons of different token pruning operations on Market-1501.}
	\centering
	\resizebox{8.2cm}{!}{
		\begin{tabular}{{l}*{2}{c}}
			\toprule
			\textbf{Methods}  & \textbf{Train Acc.} (\%) & \textbf{Val Acc.} (\%) \\
			\midrule
			ViT-B16   &  99.4  &92.4\\
			ViT-B16 + linear projection & 99.9& 60.3\\
			ViT-B16 + $1\times1$ convolution & 99.9 & 62.1\\
			ViT-B16 + depth-wise convolution     & 99.5 & 92.2 \\
			\textbf{ViT-B16 + index} & 99.4 &92.4\\
			\bottomrule	
	\end{tabular}}
	\label{tab_index}	
\end{table}

\begin{figure}[t]
	\centering
	\includegraphics[width=1\linewidth]{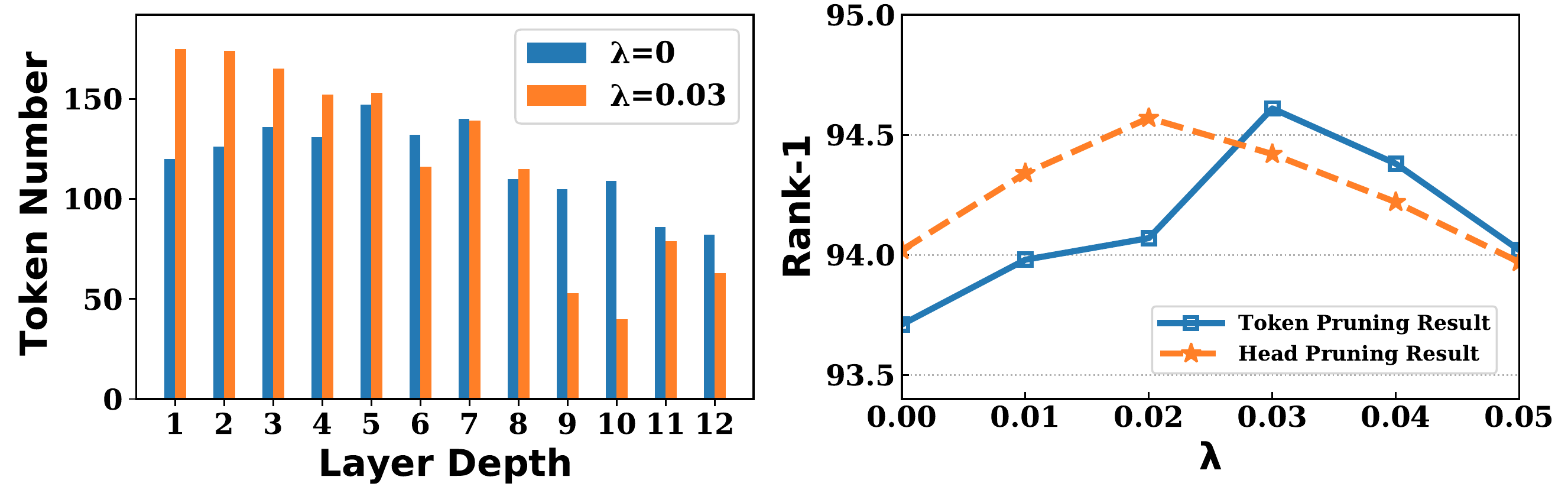}
	\caption{(a) The token structure with and without layer weighted global pruning. (b) The Rank-1 results with the change of $\lambda$ on token and head pruning.}
	\label{fig4}
\end{figure}

\paragraph{\textbf{Pruning on Small Version for Hardware Application}} Considering the powerful representation ability of ViT, we replace ViT-Base with ViT-Small as the backbone of TransReid for a better speed-accuracy trade-off. We prune the TR-Small for a further compressed model which can be easily applied on edge devices. Table~\ref{tab_trans_small} presents our pruning results on TR-Small. We prune heads and tokens at two different pruning rates respectively (25\% and 40\%). We have much better results compared with ViT-base pruning. As the results have shown, we have an improvement when 25\% heads or tokens are removed. The Rank-1 increased 0.3\% and 0.5\% while mAP grows 0\% and 0.3\% for token and head pruning. The exciting results of our pruning on 25\% verify the effectiveness of the method and encourage us to further explore the pruning extreme of head and token pruning. We prune 40\% tokens and find a marginal performance drop, 0.7\% for Rank-1 and 1.0\% for mAP. This is easy to understand because a larger pruning rate will inevitably remove discriminative tokens, which will greatly affect accuracy. We also eliminate 40\% of the heads and retain the comparable accuracy, with only a 1.0\% decline on Rank-1 and a 0.7\% increase on mAP. We further prune the Head40 model with 40\% tokens removed and get the final compressed model with only 55.7\% flops and 60\% MSA parameters. The cost is merely 0\% Rank-1 and 2.2\% mAP drop for the 1.66$\times$ speed up.

\paragraph{\textbf{Comparison to the State-of-the-arts}}
Since our method can even improve the performance with a small pruning rate, we compare our results with state-of-the-arts on both occluded and holistic datasets in Table~\ref{tab_compare_sota}. Our pruning result on Market-1501 does not improve accuracy because the model performance has achieved a high level, making it hard to improve. It is also comparable to other methods with only 0.4\% lower than the best. The performance of Occlude-Duke still has considerable room for improvement. The accuracy of TransReid in the paper~\cite{he2021transreid} is lower than the baseline in Table~\ref{tab_trans_b} because we fine-tune the original model for a few epochs to balance the training time between our pruned model and baseline. After removing 25\% of the heads, the accuracy is further improved, surpassing all other SOTA methods.

\paragraph{\textbf{Pruing on ImageNet}} To fully validate the effectiveness of our pruning methods, we also experiment with the ImageNet dataset. Table~\ref{tab_imgnet} presents the performances of our approach and other ViT compression methods on ImageNet. By observing the experiment results, our pruning methods are also effective on the large-scale dataset ImageNet. With 25\% heads or tokens pruned, the top-1 accuracy only drops by 0.1\% (removing 25\% heads on ViT-Small) and even increases (pruning 25\% tokens on ViT-Tiny and ViT-Small). Compared with other compression methods, our AMG pruning method has comparable performance. For example, the original ViT-Base model performs worse than Swin-B, with much higher FLOPs and lower accuracy. After pruning with our AMG method, the accuracy is increased to exceed Swin-B, while the FLOPs are approaching Swin-B.

\subsection{Ablation Studies}
\paragraph{\textbf{Index Layer \vs~Projection Layer}}
Memory compression methods on NLP tasks like \cite{2017exploiting,sun2022pnp} use $1\times1$ convolution or linear matrix to project key \& value tokens to a smaller number. One straightforward idea is to use these methods to prune tokens.  We implement these methods to the ViT model and the results are shown in Table~\ref{tab_index}. 
We observe that these methods have high training accuracy and extremely low test accuracy. We conjecture that the reason for this is the interaction between tokens in the projecting process. To support our conjecture, we use the depth-wise convolution to perform projection, we find that it obtains a validation accuracy of $92.2\%$. 
To this end, we propose the index layer shown in Fig.~\ref{fig2} to avoid the interaction without affecting the accuracy. Moreover, the simple design of the index layer has fewer parameters and computation added compared to the projection-based methods. Therefore, it is friendly for edge device deployment.

\paragraph{\textbf{Effectiveness of Layer Weighted Global Pruning}}
Before gallery all the scores of heads or tokens for pruning globally, we multiply them with layer coefficients according to their layer depth. The coefficients will make heads or tokens in deep layers more likely to be pruned. We take this strategy when we find the pruned structure not following the common thoughts that deeper is less important. In Fig.~\ref{fig4} (a), we can see that too many tokens are pruned in the first layer when our strategy is not adopted. The structure of our strategy is more close to the ideal winning ticket. Fig.~\ref{fig4} (b) shows the results of taking different coefficients $\lambda$ when pruning. $\lambda = 0$ indicates that our layer weighted is not adopted and a larger $\lambda$ indicates that more heads or tokens will be pruned in the deep layer. For both token and head pruning, the accuracy will firstly increase as $\lambda$ grows, this strongly verifying the effect of our strategy. When $\lambda$ continues to grow, the accuracy will start to drop since too many heads or tokens are pruned in the deep layer.

\section{Conclusions}
In this work, we propose an attention map guided pruning method on ViT for occluded person Re-ID on edge devices. We fully use the information of attention map along key and query dimensions to guide the head and token pruning respectively. We combine the two methods to extend the pruning rate further. For edge device application, we design an index layer and a dimension-aligned head pruning operation for actual compression without the need for hardware support. We also take a layer-weighted global pruning strategy for the ideal pruned structure. Extensive experiments on the Market-1501 and Occluded-DukeMTMC datasets demonstrate the superiority of our proposed approach.
\bibliographystyle{bib/IEEEtran}
\bibliography{bib/IEEEabrv,bib/IEEEreference}

\begin{thebibliography}{10}
\providecommand{\url}[1]{#1}
\csname url@samestyle\endcsname
\providecommand{\newblock}{\relax}
\providecommand{\bibinfo}[2]{#2}
\providecommand{\BIBentrySTDinterwordspacing}{\spaceskip=0pt\relax}
\providecommand{\BIBentryALTinterwordstretchfactor}{4}
\providecommand{\BIBentryALTinterwordspacing}{\spaceskip=\fontdimen2\font plus
\BIBentryALTinterwordstretchfactor\fontdimen3\font minus
  \fontdimen4\font\relax}
\providecommand{\BIBforeignlanguage}[2]{{%
\expandafter\ifx\csname l@#1\endcsname\relax
\typeout{** WARNING: IEEEtran.bst: No hyphenation pattern has been}%
\typeout{** loaded for the language `#1'. Using the pattern for}%
\typeout{** the default language instead.}%
\else
\language=\csname l@#1\endcsname
\fi
#2}}
\providecommand{\BIBdecl}{\relax}
\BIBdecl

\bibitem{dosovitskiy2020image}
A.~Dosovitskiy, L.~Beyer, A.~Kolesnikov, D.~Weissenborn, X.~Zhai,
  T.~Unterthiner, M.~Dehghani, M.~Minderer, G.~Heigold, S.~Gelly \emph{et~al.},
  ``An image is worth 16x16 words: Transformers for image recognition at
  scale,'' \emph{Proc. International Conference on Learning Representations},
  2021.

\bibitem{zhao2021incremental}
C.~Zhao, X.~Lv, S.~Dou, S.~Zhang, J.~Wu, and L.~Wang, ``Incremental generative
  occlusion adversarial suppression network for person reid,'' \emph{{IEEE}
  Trans. Image Process.}, vol.~30, pp. 4212--4224, 2021.

\bibitem{wei2018glad}
L.~Wei, S.~Zhang, H.~Yao, W.~Gao, and Q.~Tian, ``Glad: Global--local-alignment
  descriptor for scalable person re-identification,'' \emph{{IEEE} Trans.
  Multimedia}, vol.~21, no.~4, pp. 986--999, 2018.

\bibitem{luo2020stnreid}
H.~Luo, W.~Jiang, X.~Fan, and C.~Zhang, ``Stnreid: Deep convolutional networks
  with pairwise spatial transformer networks for partial person
  re-identification,'' \emph{{IEEE} Trans. Multimedia}, vol.~22, no.~11, pp.
  2905--2913, 2020.

\bibitem{zhao2020deep}
C.~Zhao, X.~Lv, Z.~Zhang, W.~Zuo, J.~Wu, and D.~Miao, ``Deep fusion feature
  representation learning with hard mining center-triplet loss for person
  re-identification,'' \emph{{IEEE} Trans. Multimedia}, vol.~22, no.~12, pp.
  3180--3195, 2020.

\bibitem{zhao2021salience}
C.~Zhao, Y.~Tu, Z.~Lai, F.~Shen, H.~T. Shen, and D.~Miao, ``Salience-guided
  iterative asymmetric mutual hashing for fast person re-identification,''
  \emph{{IEEE} Trans. Image Process.}, vol.~30, pp. 7776--7789, 2021.

\bibitem{jia2022learning}
M.~Jia, X.~Cheng, S.~Lu, and J.~Zhang, ``Learning disentangled representation
  implicitly via transformer for occluded person re-identification,''
  \emph{{IEEE} Trans. Multimedia}, 2022.

\bibitem{li2021diverse}
Y.~Li, J.~He, T.~Zhang, X.~Liu, Y.~Zhang, and F.~Wu, ``Diverse part discovery:
  Occluded person re-identification with part-aware transformer,'' in
  \emph{Proc. IEEE Conference on Computer Vision and Pattern Recognition},
  2021, pp. 2898--2907.

\bibitem{touvron2021training}
H.~Touvron, M.~Cord, M.~Douze, F.~Massa, A.~Sablayrolles, and H.~J{\'e}gou,
  ``Training data-efficient image transformers \& distillation through
  attention,'' in \emph{Proc. International Conference on Machine Learning},
  2021, pp. 10\,347--10\,357.

\bibitem{wang2016joint}
H.~Wang, T.~Tian, M.~Ma, and J.~Wu, ``Joint compression of near-duplicate
  videos,'' \emph{{IEEE} Trans. Multimedia}, vol.~19, no.~5, pp. 908--920,
  2016.

\bibitem{tan2017optimal}
B.~Tan, H.~Cui, J.~Wu, and C.~W. Chen, ``An optimal resource allocation for
  superposition coding-based hybrid digital--analog system,'' \emph{IEEE
  Internet of Things Journal}, vol.~4, no.~4, pp. 945--956, 2017.

\bibitem{gaikwad2022end}
B.~Gaikwad and A.~Karmakar, ``End-to-end person re-identification: Real-time
  video surveillance over edge-cloud environment,'' \emph{Computers and
  Electrical Engineering}, vol.~99, p. 107824, 2022.

\bibitem{camara2020pedestrian}
F.~Camara, N.~Bellotto, S.~Cosar, D.~Nathanael, M.~Althoff, J.~Wu, J.~Ruenz,
  A.~Dietrich, and C.~W. Fox, ``Pedestrian models for autonomous driving part
  i: low-level models, from sensing to tracking,'' \emph{IEEE Transactions on
  Intelligent Transportation Systems}, vol.~22, no.~10, pp. 6131--6151, 2020.

\bibitem{molchanov2016pruning}
P.~Molchanov, S.~Tyree, T.~Karras, T.~Aila, and J.~Kautz, ``Pruning
  convolutional neural networks for resource efficient inference,'' \emph{Proc.
  International Conference on Learning Representations}, 2017.

\bibitem{molchanov2019importance}
P.~Molchanov, A.~Mallya, S.~Tyree, I.~Frosio, and J.~Kautz, ``Importance
  estimation for neural network pruning,'' in \emph{Proc. IEEE Conference on
  Computer Vision and Pattern Recognition}, 2019, pp. 11\,264--11\,272.

\bibitem{chen2020lottery}
T.~Chen, J.~Frankle, S.~Chang, S.~Liu, Y.~Zhang, Z.~Wang, and M.~Carbin, ``The
  lottery ticket hypothesis for pre-trained bert networks,'' \emph{Advances in
  Neural Information Processing Systems}, 2020.

\bibitem{tang2021patch}
Y.~Tang, K.~Han, Y.~Wang, C.~Xu, J.~Guo, C.~Xu, and D.~Tao, ``Patch slimming
  for efficient vision transformers,'' \emph{arXiv preprint arXiv:2106.02852},
  2021.

\bibitem{wang2021pyramid}
W.~Wang, E.~Xie, X.~Li, D.-P. Fan, K.~Song, D.~Liang, T.~Lu, P.~Luo, and
  L.~Shao, ``Pyramid vision transformer: A versatile backbone for dense
  prediction without convolutions,'' \emph{Proc. IEEE International Conference
  on Computer Vision}, pp. 568--578, 2021.

\bibitem{rethinking2021}
B.~Heo, S.~Yun, D.~Han, S.~Chun, J.~Choe, and S.~J. Oh, ``Rethinking spatial
  dimensions of vision transformers,'' \emph{Proc. IEEE International
  Conference on Computer Vision}, pp. 11\,936--11\,945, 2021.

\bibitem{xu2021evo}
Y.~Xu, Z.~Zhang, M.~Zhang, K.~Sheng, K.~Li, W.~Dong, L.~Zhang, C.~Xu, and
  X.~Sun, ``Evo-vit: Slow-fast token evolution for dynamic vision
  transformer,'' \emph{arXiv preprint arXiv:2108.01390}, 2021.

\bibitem{chen2021chasing}
T.~Chen, Y.~Cheng, Z.~Gan, L.~Yuan, L.~Zhang, and Z.~Wang, ``Chasing sparsity
  in vision transformers: An end-to-end exploration,'' \emph{Advances in Neural
  Information Processing Systems}, 2021.

\bibitem{yu2022width}
F.~Yu, K.~Huang, M.~Wang, Y.~Cheng, W.~Chu, and L.~Cui, ``Width \& depth
  pruning for vision transformers,'' in \emph{Proc. AAAI Conference on
  Artificial Intelligence}, vol. 2022, 2022.

\bibitem{yu2022unified}
S.~Yu, T.~Chen, J.~Shen, H.~Yuan, J.~Tan, S.~Yang, J.~Liu, and Z.~Wang,
  ``Unified visual transformer compression,'' \emph{Proc. International
  Conference on Learning Representations}, 2022.

\bibitem{zhuo2018occluded}
J.~Zhuo, Z.~Chen, J.~Lai, and G.~Wang, ``Occluded person re-identification,''
  in \emph{Proc. IEEE International Conference on Multimedia and Expo}.\hskip
  1em plus 0.5em minus 0.4em\relax IEEE, 2018, pp. 1--6.

\bibitem{he2020guided}
L.~He and W.~Liu, ``Guided saliency feature learning for person
  re-identification in crowded scenes,'' in \emph{Proc. European Conference on
  Computer Vision}.\hskip 1em plus 0.5em minus 0.4em\relax Springer, 2020, pp.
  357--373.

\bibitem{he2019foreground}
L.~He, Y.~Wang, W.~Liu, H.~Zhao, Z.~Sun, and J.~Feng, ``Foreground-aware
  pyramid reconstruction for alignment-free occluded person
  re-identification,'' in \emph{Proc. IEEE International Conference on Computer
  Vision}, 2019, pp. 8450--8459.

\bibitem{huang2020human}
H.~Huang, X.~Chen, and K.~Huang, ``Human parsing based alignment with
  multi-task learning for occluded person re-identification,'' in \emph{Proc.
  IEEE International Conference on Multimedia and Expo}.\hskip 1em plus 0.5em
  minus 0.4em\relax IEEE, 2020, pp. 1--6.

\bibitem{yu2021neighbourhood}
S.~Yu, D.~Chen, R.~Zhao, H.~Chen, and Y.~Qiao, ``Neighbourhood-guided feature
  reconstruction for occluded person re-identification,'' \emph{arXiv preprint
  arXiv:2105.07345}, 2021.

\bibitem{wang2022feature}
Z.~Wang, F.~Zhu, S.~Tang, R.~Zhao, L.~He, and J.~Song, ``Feature erasing and
  diffusion network for occluded person re-identification,'' in \emph{Proc.
  IEEE Conference on Computer Vision and Pattern Recognition}, 2022, pp.
  4754--4763.

\bibitem{he2021transreid}
S.~He, H.~Luo, P.~Wang, F.~Wang, H.~Li, and W.~Jiang, ``Transreid:
  Transformer-based object re-identification,'' in \emph{Proc. IEEE Conference
  on Computer Vision and Pattern Recognition}, 2021, pp. 15\,013--15\,022.

\bibitem{zhu2021visual}
M.~Zhu, K.~Han, Y.~Tang, and Y.~Wang, ``Visual transformer pruning,''
  \emph{arXiv preprint arXiv:2104.08500}, 2021.

\bibitem{vaswani2017attention}
A.~Vaswani, N.~Shazeer, N.~Parmar, J.~Uszkoreit, L.~Jones, A.~N. Gomez,
  {\L}.~Kaiser, and I.~Polosukhin, ``Attention is all you need,'' in
  \emph{Advances in Neural Information Processing Systems}, 2017, pp.
  5998--6008.

\bibitem{li2016pruning}
H.~Li, A.~Kadav, I.~Durdanovic, H.~Samet, and H.~P. Graf, ``Pruning filters for
  efficient convnets,'' \emph{Proc. International Conference on Learning
  Representations}, 2017.

\bibitem{2021non}
Y.~Yao, T.~Chen, G.-S. Xie, C.~Zhang, F.~Shen, Q.~Wu, Z.~Tang, and J.~Zhang,
  ``Non-salient region object mining for weakly supervised semantic
  segmentation,'' in \emph{CVPR}, 2021, pp. 2623--2632.

\bibitem{2021jo}
Y.~Yao, Z.~Sun, C.~Zhang, F.~Shen, Q.~Wu, J.~Zhang, and Z.~Tang, ``Jo-src: A
  contrastive approach for combating noisy labels,'' in \emph{CVPR}, 2021, pp.
  5192--5201.

\bibitem{pei2022eccv}
G.~Pei, F.~Shen, Y.~Yao, G.-S. Xie, Z.~Tang, and J.~Tang, ``Hierarchical
  feature alignment network for unsupervised video object segmentation,'' in
  \emph{Proc. European Conference on Computer Vision}, 2022, pp. 596--613.

\bibitem{rao2021dynamicvit}
Y.~Rao, W.~Zhao, B.~Liu, J.~Lu, J.~Zhou, and C.-J. Hsieh, ``Dynamicvit:
  Efficient vision transformers with dynamic token sparsification,''
  \emph{Advances in Neural Information Processing Systems}, 2021.

\bibitem{liao2021searching}
Y.-L. Liao, S.~Karaman, and V.~Sze, ``Searching for efficient multi-stage
  vision transformers,'' \emph{Proc. IEEE International Conference on Computer
  Vision}, 2021.

\bibitem{liu2021swin}
Z.~Liu, Y.~Lin, Y.~Cao, H.~Hu, Y.~Wei, Z.~Zhang, S.~Lin, and B.~Guo, ``Swin
  transformer: Hierarchical vision transformer using shifted windows,''
  \emph{Proc. IEEE International Conference on Computer Vision}, pp.
  10\,012--10\,022, 2021.

\bibitem{gao2020pose}
S.~Gao, J.~Wang, H.~Lu, and Z.~Liu, ``Pose-guided visible part matching for
  occluded person reid,'' in \emph{Proc. IEEE Conference on Computer Vision and
  Pattern Recognition}, 2020, pp. 11\,744--11\,752.

\bibitem{miao2019pose}
J.~Miao, Y.~Wu, P.~Liu, Y.~Ding, and Y.~Yang, ``Pose-guided feature alignment
  for occluded person re-identification,'' in \emph{Proc. IEEE International
  Conference on Computer Vision}, 2019, pp. 542--551.

\bibitem{wang2020high}
G.~Wang, S.~Yang, H.~Liu, Z.~Wang, Y.~Yang, S.~Wang, G.~Yu, E.~Zhou, and
  J.~Sun, ``High-order information matters: Learning relation and topology for
  occluded person re-identification,'' in \emph{Proc. IEEE Conference on
  Computer Vision and Pattern Recognition}, 2020, pp. 6449--6458.

\bibitem{chen2021occlude}
P.~Chen, W.~Liu, P.~Dai, J.~Liu, Q.~Ye, M.~Xu, Q.~Chen, and R.~Ji, ``Occlude
  them all: Occlusion-aware attention network for occluded person re-id,'' in
  \emph{Proc. IEEE Conference on Computer Vision and Pattern Recognition},
  2021, pp. 11\,833--11\,842.

\bibitem{sun2018beyond}
Y.~Sun, L.~Zheng, Y.~Yang, Q.~Tian, and S.~Wang, ``Beyond part models: Person
  retrieval with refined part pooling (and a strong convolutional baseline),''
  in \emph{Proc. European Conference on Computer Vision}, 2018, pp. 480--496.

\bibitem{luo2019bag}
H.~Luo, Y.~Gu, X.~Liao, S.~Lai, and W.~Jiang, ``Bag of tricks and a strong
  baseline for deep person re-identification,'' in \emph{Proc. IEEE Conference
  on Computer Vision and Pattern Recognition}, 2019, pp. 0--0.

\bibitem{zheng2015scalable}
L.~Zheng, L.~Shen, L.~Tian, S.~Wang, J.~Wang, and Q.~Tian, ``Scalable person
  re-identification: A benchmark,'' in \emph{Proc. IEEE International
  Conference on Computer Vision}, 2015, pp. 1116--1124.

\bibitem{deng2009imagenet}
J.~Deng, W.~Dong, R.~Socher, L.-J. Li, K.~Li, and L.~Fei-Fei, ``Imagenet: A
  large-scale hierarchical image database,'' in \emph{Proc. IEEE Conference on
  Computer Vision and Pattern Recognition}, 2009, pp. 248--255.

\bibitem{2017exploiting}
Y.~Yao, J.~Zhang, F.~Shen, X.~Hua, J.~Xu, and Z.~Tang, ``Exploiting web images
  for dataset construction: A domain robust approach,'' \emph{TMM}, vol.~19,
  no.~8, pp. 1771--1784, 2017.

\bibitem{sun2022pnp}
Z.~Sun, F.~Shen, D.~Huang, Q.~Wang, X.~Shu, Y.~Yao, and J.~Tang, ``Pnp: Robust
  learning from noisy labels by probabilistic noise prediction,'' in
  \emph{Proc. IEEE Conference on Computer Vision and Pattern Recognition},
  2022, pp. 5311--5320.

\end{thebibliography}

\end{document}